# A GFML-based Robot Agent for Human and Machine Cooperative Learning on Game of Go

Chang-Shing Lee, Mei-Hui Wang
Li-Chuang Chen
National University of Tainan
Tainan, Taiwan
leecs@mail.nutn.edu.tw
mh.alice.wang@gmail.com
gh1321223@gmail.com

Yusuke Nojima
Osaka Prefecture University
Osaka, Japan
nojima@cs.osakafu-u.ac.jp

Tzong-Xiang Huang, Jinseok Woo
Naoyuki Kubota, Eri Sato-Shimokawara, Toru Yamaguchi
Tokyo Metropolitan University
Tokyo, Japan
huang-tzong-xiang@ed.tmu.ac.jp
woojs@tmu.ac.jp, kubota@tmu.ac.jp
eri@tmu.ac.jp, yamachan@tmu.ac.jp

*Abstract*—**This paper applies a genetic algorithm and fuzzy markup language to construct a human and smart machine cooperative learning system on game of Go. The genetic fuzzy markup language (GFML)-based Robot Agent can work on various kinds of robots, including Palro, Pepper, and TMU's robots. We use the parameters of FAIR open source Darkforest and OpenGo AI bots to construct the knowledge base of Open Go Darkforest (OGD) cloud platform for student learning on the Internet. In addition, we adopt the data from AlphaGo Master's sixty online games as the training data to construct the knowledge base and rule base of the co-learning system. First, the Darkforest predicts the win rate based on various simulation numbers and matching rates for each game on OGD platform, then the win rate of OpenGo is as the final desired output. The experimental results show that the proposed approach can improve knowledge base and rule base of the prediction ability based on Darkforest and OpenGo AI bot with various simulation numbers.**

## I. INTRODUCTION

The current robotic system based on Facebook ELF OpenGo [9, 10] can provide the five best moves to a player. From the viewpoint of education, the robot should recommend the five best moves at the appropriate timing when the player really needs any hint. It means that the robot should not give the recommendation to the player every move. If the robot tells the best move to the player every turn, the player will seldom think the situation deeply and maybe lose the opportunity to improve his/her play skill easily. The kind of information (i.e., recommendation) is also important. The recommendation of the five best moves could be very useful for the player. But it may be not true for every time. Assume the following situation. Now, the player is wondering which move he/she should choose from a few best candidates. If the best move suggested by the robot is the same as one of candidate moves that the player is thinking, the recommendation convinces the player to choose the best move. If the five best moves are different from the player's thought, the player may be able to find why the recommended best moves are better under his/her experience. However, if the player does not have enough experience, he/she cannot understand why the recommended best moves are better options. At that situation, the player needs the reason to improve his/her play skill. The above discussions indicate that the timing and kind of information should be considered for providing the meaningful cooperation between a human and a robot.

The main aim of fuzzy markup language (FML), a XML-based language, is to bridge the aforementioned implementation gaps by introducing an abstract and unified approach for designing fuzzy systems in hardware independent way [1]. FML is a specific-purpose computer language and allows systems' designers to express their ideas in fast and simple way and speeds up the whole development process of a given complex system [2]. FML has been the first IEEE standard (IEEE Std 1855-2016) sponsored by the IEEE Computational Intelligence Society (CIS) [3, 4]. JFML is an open source Java library which is aimed at facilitating interoperability and usability of fuzzy systems [5] and provides some examples of applications like diet assessment, and tipper. Lee et al. also used FML to construct knowledge base and rule base of the fuzzy system related to applications to student learning performance [6], game of Go [7], Japanese dietary assessment [8], and so on.

With the success of AlphaGo [11, 12], there has been a lot of interest among students and professionals to apply machine learning to gaming and in particular to the game of Go. Several conferences have held competitions human vs. computer programs or computer programs against each other. While computer programs are already better than humans (even high level professionals), machine learning still offers interesting prospects, both from the fundamental point of view: 1) to even further the limits of game playing (having programs playing against each other), and 2) to better understand machine intelligence and compare it to human intelligence, and from the practical point of view of enhancing the human playing experience by coaching professionals to play better or training beginners. The latter problem raises interesting questions of explainability of machine game play.

An FML-based machine learning competition for human and smart machine co-learning on game of Go will be held in IEEE CEC 2019 conference to evaluate the potential of learning machines to teach humans [13]. This paper uses the data from AlphaGo Master's sixty online games as the training data to

The authors would like to thank the financially support sponsored by the Ministry of Science and Technology of Taiwan under the grants MOST 107-2218-E-024-001.

construct the knowledge base and rule base of the co-learning system. We hope to optimize the parameters of the constructed knowledge base and rule base after learning the ones of FAIR OpenGo to evaluate the potential of learning machines to teach humans. The remainder of this paper is organized as follows: Sections II and III introduce the human and robot cooperative learning system and GFML robot agent, respectively. The experimental results are shown in Section IV and conclusions are given in Section V.

## II. HUMAN AND ROBOT COOPERATIVE LEARNING SYSTEM

### A. Smartphone Interlocked Robot Partners

In this section, we introduce the robot partner system using smart devices. Smart devices are equipped with various sensors such as gyro, accelerometer, illumination sensor, touch interface, compass, two cameras, and microphone. The configuration of sufficient sensors is necessary to construct a robot partner. Additionally, the robot partner is equipped with servomotors on the robot body for its movement based on a microcontroller. Therefore, our robot partner has been able to develop robots with various interfaces based on the combination of smart device services and hardware for various purposes as shown in Fig. 1 [14-17]. For example, Fig. 1(a) shows that we can design various styles of the robot partner "iPhonoid" to meet users' needs only using a basic smart device [14-15]. Fig. 1(b) shows the robot partner "iPadrone" which is mainly used for a human-robot interaction through the equipped touch screen [16]. Fig. 1(c) shows the robot partner "concierge" and it could be used in shopping malls, event halls, public offices, and other places for multi-language support [18].

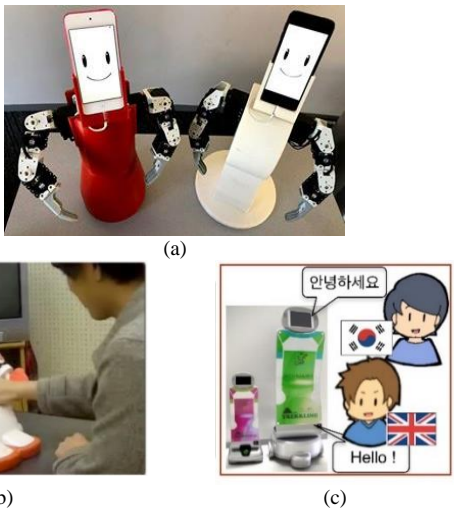

Fig. 1. Smartphone interlocked robot partners, including (a) iPhonoid series (red: iPhonoid-C and white: iPhonoid-D), (b) iPadrone, and (c) concierge robot [14-18].

### B. The Timing of Suggestions by a Robot

The timing of suggestions by a robot is very important for human and smart machine co-learning system. If the robot gives suggestions every turn, the player may stop thinking by his/herself and just follow the most promising suggestion. This loses the player's learning opportunity. In order to avoid this situation, the robot should detect an appropriate timing to give suggestions to the player. 1) **Option 1- Elapsed thinking time**: If the player is thinking the next move for a long time, it may be a good time to give suggestions to the player. A difficulty is how to specify an appropriate threshold on the elapsed thinking time. 2) **Option 2- Brain condition**: If it is possible to observe the player's mind state using BMI, it may be possible to determine the timing to give suggestions to the player. A difficulty is how to define the mind state which indicates that the player needs any hint. 3) **Option 3- Turning point**: If it is possible to detect turning points of the game by the robot, it may be a good time to tell something to the player. The exact best move may be not necessary.

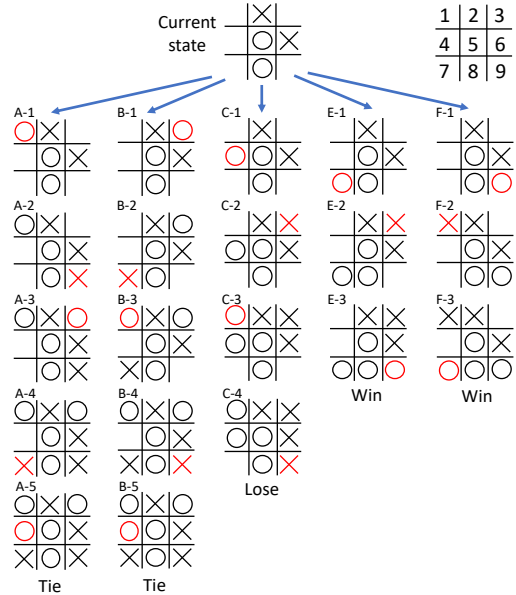

Fig. 2. Tic-tac-toe.

The kind of information by a robot is also very important. It should be considered that the suggestions by the robot are not always understandable for any player. It is totally depended on the player and situation. Suggesting only five best moves is sometimes good for the player to confirm whether the next move he/she thinks beforehand is a good choice or not. Even if the suggested moves are different from the player's thought, it may be a good moment to consider other possible moves. However, if the suggested moves are far different from the player's thought, the player cannot understand why those moves are better in the facing situation. This may lead to the player's confusion. From the educational point of view, the player should know not only the best moves but also the reason why those moves are best. The robot should provide additional information or the reason to make the player imagine the future game state if the player chooses one of the best moves. For example, assume the current state in the tic-tac-toe game of Fig. 2. You are circle and thinking the next move. There are five possible positions for you: 1, 3, 4, 7, 9. The best moves are 7 or 9. It is trivial because you can easily imagine the future states like E-3 or F-3. However, if you cannot imagine the next states even like E-2 or F-2, you cannot understand why the move 7 or 9 are the best. In such a case, showing the future state like E-2, F-2, E-3, or F-3 helps the player to take a decision or to understand why the best moves change the future state. The robot should not always give the best moves. The robot should make the player think the next move and show how the future states will change under the player's decision. That is, when the player is about to choose the move 4, the robot shows the future states like C-2 or C-5. This gives the chance to deeply think the next move again.

### C. Future State Prediction

One simple idea to implement the above idea is to perform an internal simulation (like mental simulation in human brain). Fig. 3 illustrates the idea. The current robot provides only the top five moves (i.e. blue arrow in Fig. 3) without any explanation. The new architecture temporarily choose one move (i.e. red arrow in Fig. 3) and send it to the OGD cloud platform. The OGD cloud platform provides the next top five moves. Then one of the moves is temporarily chosen again. This process iterates several times. For each temporal choice, the robot can provide a promising future state (i.e. green arrow) to the player. According to the prediction of the future states, the player can imagine how the game will change by the player's next move.

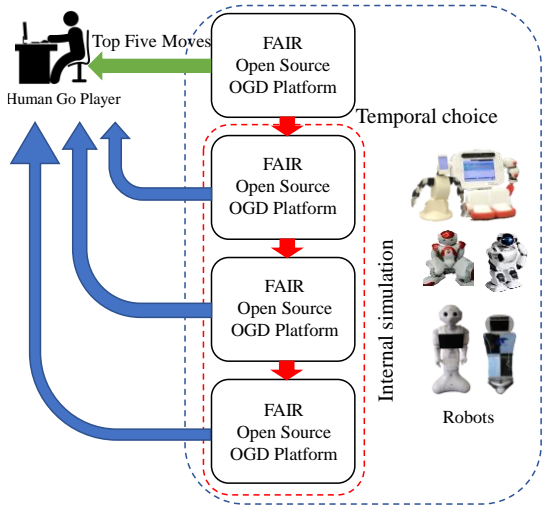

Fig. 3. Future state prediction.

## III. GFML-BASED ROBOT AGENT

### A. Structure of Human and Robot Cooperative Learning System

In this paper, we apply Genetic Fuzzy Markup Language (GFML) [20, 21] to construct a robot agent for human and smart machine cooperative learning system on game of Go. The knowledge base and rule base refer to the parameters of FAIR Darkforest and ELF OpenGo open source. Fig. 4 shows the various robots used to integrate the OGD cloud platform for the robot agent and we give some descriptions as follows:

1) The Go player surfs on the OGD Cloud Platform to play Go. The robot Palro reports the next suggested moves and current game situation to the Go player. In addition, the robot Pepper shows the predicted information or human's psychological indicators during playing. The other robots of Kubota Lab. also show the related information on their monitor at the same time.

2) The proposed robot agent is composed of a data

preprocessing module, a prediction module, a GFML-based robot learning module, a machine-based game database, and a knowledge base/rule base.

3) The predication module, including an ELF-based prediction model and a DDF-based prediction model, is responsible for predicting the position, win rate, and simulation numbers of next five moves. The predicted information is stored in the machine-based game database.

4) The *FML engineer and domain expert* construct the knowledge base and rule base of the robot agent by referring to the predicted data of the OGD cloud platform. Additionally, the *Data engineer and domain expert* also organize the DDF data and ELF data for the data preprocessing module to select the training data and testing data.

5) The GFML-based robot learning module is responsible for learning the parameters of the knowledge base and rule base until termination based on the preprocessed data and the machine-based game data.

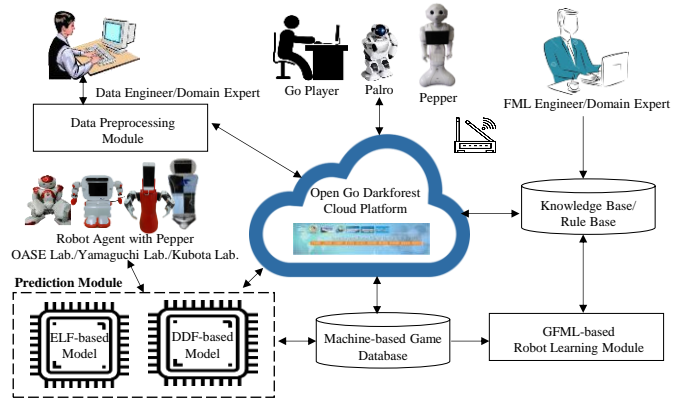

Fig. 4. Human and robot cooperative learning system structure.

### B. Data Preprocessing Module

This section introduces the data preprocessing module. First, we adopt the sixty games of Google Deepmind AlphaGo Master Go from the Internet, then apply the FAIR Darkforest Go AI bot on the OGD cloud platform to predict the win rate with various DDF simulation numbers. Second, the ELF OpenGo AI bot is utilized to compute the final ELF win rate for each game, and the ELF win rate is as the desired output for each game. Third, we define three fuzzy variables for *Black* and *White*, including, DDF win rate, DDF simulation numbers, and DDF matching rate. For the fuzzy variable *Simulation Number*, we normalize the value to the interval [0, 1]. Finally, the data from Master Game 1 to Game 45 are set to be the training data, and the data of Game 46 to Game 60 are set to be the testing data.

### C. Knowledge Base and Rule Base for the Robot Agent

There are six input fuzzy variables defined in the knowledge base, including *DDF Black Simulation Number* (*DBSN*), *DDF White Simulation Number* (*DWSN*), *DDF Black Win Rate* (*DBWR*), *DDF White Win Rate* (*DWWR*), *DDF Black Top-Move Rate* (*DBTMR*), and *DDF White Top-Move Rate* (*DWTMR*). The output fuzzy variable is *ELF Win Rate* (*EWR*). The fuzzy variables *DBSN*, *DWSN*, *DBTMR*, and *DWTMR* contain two fuzzy numbers, including *High* and *Low*, respectively. *DBWR* and *DWWR* contain three fuzzy numbers, including *High*, *Medium* and *Low*, respectively. The output fuzzy variable *EWR* contains four fuzzy numbers, including *High*, *Medium_High*, *Medium_Low*, and *Low*. Table I shows the parameters of the fuzzy variables and fuzzy numbers and Table II shows partial rule base. Table III shows the partial knowledge base and rule base represented by GFML.

TABLE I. PARAMETERS OF FUZZY VARIABLES AND FUZZY NUMBERS.

| DBSN | | DWSN | |
|---|---|---|---|
| Low | [0,0,0.4,0.6] | Low | [0,0,0.4,0.6] |
| High | [0.4,0.6,1,1] | High | [0.4,0.6,1,1] |
| DBWR | | DWWR | |
| Low | [0,0,0.3,0.4] | Low | [0,0,0.3,0.4] |
| Medium | [0.3,0.4,0.6,0.7] | Medium | [0.3,0.4,0.6,0.7] |
| High | [0.6,0.7,1,1] | High | [0.6,0.7,1,1] |
| DBTMR | | DWTMR | |
| Low | [-1,-1,-0.2,0.2] | Low | [-1,-1,-0.2,0.2] |
| High | [-0.2,0.2,1,1] | High | [-0.2,0.2,1,1] |
| EBWR / EWWR | | | |
| Low | [0,0,0.2,0.3] | | |
| Medium_Low | [0.2,0.3,0.4,0.55] | | |
| Medium_High | [0.4,0.55,0.7,0.8] | | |
| High | [0.7,0.8,1,1] | | |

TABLE II. PARTIAL RULE BASE.

| No | Fuzzy Variables | | | | | | |
|---|---|---|---|---|---|---|---|
| | DBSN | DWSN | DBWR | DWWR | DBTMR | DWTMR | EWR |
| 1 | High | High | High | High | High | High | High |
| 2 | Low | High | High | High | High | High | High |
| 3 | High | Low | High | High | High | High | High |
| 4 | Low | Low | High | High | High | High | High |
| 5 | High | High | High | High | High | High | High |
| 6 | Low | High | High | High | High | High | High |
| 7 | High | Low | High | High | High | High | High |
| 8 | Low | Low | High | High | High | High | High |
| 9 | High | High | High | High | High | High | High |
| 10 | Low | High | High | High | High | High | High |
| | | | | ⋮ | | | |
| 139 | Low | Medium | Low | Low | Low | Low | Low |
| 140 | Low | Medium | Low | Low | Low | Low | Low |
| 141 | High | Low | Low | Low | Low | Low | Low |
| 142 | High | Low | Low | Low | Low | Low | Low |
| 143 | Low | Low | Low | Low | Low | Low | Low |
| 144 | Low | Low | Low | Low | Low | Low | Low |

TABLE III. PARTIAL KNOWLEDGE BASE AND RULE BASE REPRESENTED BY GFML.

```xml
<?xml version="1.0" ?>
<FuzzyController ip="localhost" name="">
    <KnowledgeBase>
        <FuzzyVariable domainleft="0" domainright="1" name="DBSN"
scale="" type="input">
        </FuzzyTerm>
        <FuzzyTerm name="Low" hedge="Normal">
            <TrapezoidShape  Param1="0"  Param2="0"  Param3="0.4"
Param4="0.6" />
        </FuzzyTerm>
        <FuzzyTerm name="High" hedge="Normal">
            <TrapezoidShape  Param1="0.4"  Param2="0.6"  Param3="1"
Param4="1" />
        </FuzzyTerm>
        <FuzzyVariable  domainleft="0"  domainright="1"  name="DWSN"
scale="" type="input">
```

```xml
        <FuzzyTerm name="Low" hedge="Normal">
            <TrapezoidShape  Param1="0"  Param2="0"  Param3="0.4"
Param4="0.6" />
        </FuzzyTerm>
        <FuzzyTerm name="High" hedge="Normal">
            <TrapezoidShape  Param1="0.4"  Param2="0.6"  Param3="1"
Param4="1" />
        </FuzzyTerm>
    </FuzzyVariable>
                            ⋮
    </KnowledgeBase>
    <RuleBase activationMethod="MIN" andMethod="MIN"
orMethod="MAX" name="RuleBase1" type="mamdani">
        <Rule name="Rule1" connector="and" weight="1" operator="MIN">
            <Antecedent>
                <Clause>
                    <Variable>DBSN</Variable>
                    <Term>High</Term>
                </Clause>
                <Clause>
                    <Variable>DWSN</Variable>
                    <Term>High</Term>
                </Clause>
                <Clause>
                    <Variable>DBWR</Variable>
                    <Term>High</Term>
                </Clause>
                <Clause>
                    <Variable>DWWR</Variable>
                    <Term>High</Term>
                </Clause>
                <Clause>
                    <Variable>DBTMR</Variable>
                    <Term>High</Term>
                </Clause>
                <Clause>
                    <Variable>DWTMR</Variable>
                    <Term>High</Term>
                </Clause>
            </Antecedent>
            <Consequent>
                <Clause>
                    <Variable>EWR</Variable>
                    <Term>High</Term>
                </Clause>
            </Consequent>
                            ⋮
        </Rule>
    </RuleBase>
</FuzzyController>
```

## IV. EXPERIMENTAL RESULTS

### A. Robot Agent on various Robots

The proposed robot agent can deploy on various communication robots such as Palro, Nao, Pepper, and academic TMU robot partners, including iPhonoid and iPadrone. Fig. 5 shows the environment setup for testing the stability of the cooperative learning system in Jan. 2019, at TMU, Japan. We setup the developed OGD cloud platform, EFL OpenGo AI bot, and Darkforest AI bot in NCHC/Taiwan, KWS Center/Taiwan, and Nojim Lab./OPU/Japan. The robots are located in Kubota Lab./ Yamaguchi Lab./TMU/Japan. Fig. 6(a) shows the invited Go player Yi-Hsiu Lee (8P) cooperated with various robots for tesing the stability of the cooperative learning system on Jan. 17, 2019 at TMU. Fig. 6(b) shows the testing scene of the Go players, including Yi-Hsiu Lee (8P) and Hirofumi Ohashi (6P). The robot pepper showed the Go players' psychological indicators. The robots iPhonoid-C and iPadrone showed the information of suggestion moves and win rate, respectively. The robot Palro reported the top suggestion move to Lee (8P) and Ohashi (6P) for a reference. The experimental results show the robot agent on various robots can work well.

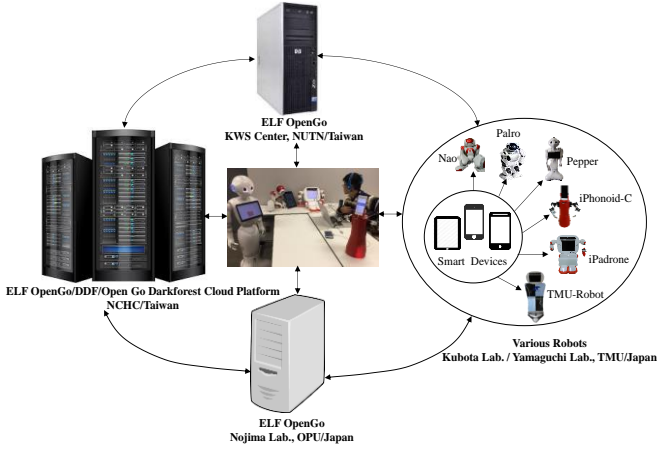

Fig. 5. Environment setup for testing the stability of the cooperative learning system.

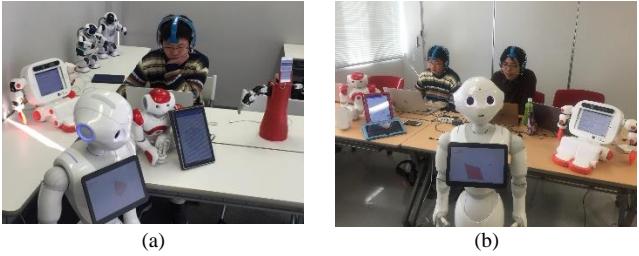

Fig. 6. Testing scenes of (a) Lee (8P) and (b) Lee (8P) & Ohashi (6P).

## B. GFML for Knwledge Base and Rule Base Tuning

To evaluate the performance of the proposed approach, we used sixty games between AlphaGo Master and professional Go players in Dec. 2016 and Jan. 2017 [19] as the experimental data. The first forty-five games with 4178 records are the training data and the last fifty games with 1428 records are the testing data. The proposed GFML-based robot learning module uses genetic algorithm based on FML for human and machine cooperative learning system on game of Go. The crossover rate is set to be 0.9, the mutation rate is set to be 0.1, and the generations is set to be 10000. The desired output is set to be the win rate predicted by ELF OpenGo. Table IV shows partial information of games between AlphaGo Master and professional Go players [19]. Table V shows partial experimental training data. The learned knowledge bases for *Black* and *White* are shown in Fig. 7 and Fig. 8, respectively, after learning for 10000 generations. Figs. 9 and 10 show the *Black* and *White's* win rate curves of DBWR/DWWR predicted by Darkforest, EBWR/EWWR (DO) predicted by ELF OpenGo, and after-learning EBWR/EWWR (AL) for Games 11 and 46, respectively. Figs. 7 and 8 indicate that DBWR/DWWR has a tendency to approach to EBWR(DO)/EWWR(DO) after learning.

TABLE IV. PARTIAL INFORMATION OF ALPHAGO MASTER VS. PROFESSIONAL GO PLAYERS GAMES.

| No | Date | Black | White | TotalMove |
|---|---|---|---|---|
| 1 | 12/29/2016 | Pan Tingyu | Master | 146 |
| 2 | 12/29/2016 | Zhang Ziliang | Master | 174 |
| 3 | 12/29/2016 | Master | Ding Shixiong | 151 |
| 4 | 12/29/2016 | Master | Xie Erhao | 222 |
| 5 | 12/29/2016 | Master | Yu Zhiying | 113 |
| 6 | 12/29/2016 | Master | Li Xiangyu | 131 |
| 7 | 12/29/2016 | Master | Qiao Zhijian | 163 |
| 8 | 12/29/2016 | Qiao Zhijian | | 104 |
| 9 | 12/29/2016 | | Meng Tailing | 275 |
| 10 | 12/29/2016 | Meng Tailing | Master | 148 |
| ⋮ | | | | |
| 51 | 1/4/2017 | Zhou Junxun | Master | 118 |
| 52 | 1/4/2017 | Fan Tingyu | Master | 202 |
| 53 | 1/4/2017 | Master | Huang Yunsong | 133 |
| 54 | 1/4/2017 | Master | Nie Weiping | 254 |
| 55 | 1/4/2017 | Chen Yaoye & Meng Tailing | Master | 267 |
| 56 | 1/4/2017 | Master | Cho Hanseung | 171 |
| 57 | 1/4/2017 | Master | Shin Jinseo | 139 |
| 58 | 1/4/2017 | Chang Hao | Master | 178 |
| 59 | 1/4/2017 | Master | Zhou Ruiyang | 161 |
| 60 | 1/4/2017 | Gu Li | Master | 235 |

TABLE V. PARTIAL EXPERIMENTAL TRAINING DATA.

| No | DBSN | DWSN | DBWR | DWWR | DBTMR | DWTMR | EBWR | EWWR |
|---|---|---|---|---|---|---|---|---|
| 1 | 0.3863 | 0.2274 | 0.52733 | 0.48533 | 0 | 1 | 0.49636 | 0.50884 |
| 2 | 0.9283 | 0.7866 | 0.51529 | 0.48532 | 0.5 | 1 | 0.45779 | 0.54844 |
| 3 | 1 | 0.6798 | 0.51265 | 0.47717 | 0.6667 | 1 | 0.45671 | 0.57235 |
| 4 | 0.4499 | 1 | 0.51885 | 0.46988 | 0.75 | 1 | 0.49086 | 0.5201 |
| 5 | 0.7388 | 1 | 0.5288 | 0.46679 | 0.8 | 1 | 0.47218 | 0.53649 |
| 6 | 1 | 0.9693 | 0.53309 | 0.46602 | 0.8333 | 1 | 0.47249 | 0.52224 |
| 7 | 1 | 1 | 0.53075 | 0.4713 | 0.8571 | 1 | 0.4811 | 0.49862 |
| 8 | 0.6786 | 0.4892 | 0.5212 | 0.45576 | 0.875 | 0.875 | 0.48858 | 0.53774 |
| 9 | 1 | 0.6432 | 0.53276 | 0.46472 | 0.8889 | 0.7778 | 0.48073 | 0.53004 |
| 10 | 0.5267 | 1 | 0.51407 | 0.4886 | 0.9 | 0.8 | 0.486 | 0.50764 |
| 11 | 0.8314 | 1 | 0.50868 | 0.48669 | 0.9091 | 0.8182 | 0.43671 | 0.58388 |
| 12 | 1 | 1 | 0.51329 | 0.48678 | 0.9167 | 0.8333 | 0.41195 | 0.59656 |
| 13 | 1 | 0.6694 | 0.51694 | 0.49295 | 0.9231 | 0.8462 | 0.40338 | 0.59961 |
| 14 | 0.6409 | 1 | 0.50828 | 0.5085 | 0.9286 | 0.8571 | 0.34558 | 0.70091 |
| 15 | 1 | 1 | 0.49284 | 0.50715 | 0.9333 | 0.8667 | 0.29436 | 0.71018 |
| ⋮ | | | | | | | | |
| 4174 | 1 | 0.8421 | 0.87858 | 0.1324 | 0.7848 | 0.7595 | 0.99456 | 0.00877 |
| 4175 | 0.9206 | 0.4521 | 0.8759 | 0.11748 | 0.7875 | 0.7625 | 0.99581 | 0.00592 |
| 4176 | 0.4137 | 1 | 0.89287 | 0.11602 | 0.7901 | 0.7654 | 0.99314 | 0.01174 |
| 4177 | 1 | 0.9354 | 0.88357 | 0.1207 | 0.7927 | 0.7683 | 0.99353 | 0.00774 |
| 4178 | 1 | 1 | 0.88463 | 0.11548 | 0.7952 | 0.7711 | 0.99386 | 0.00712 |

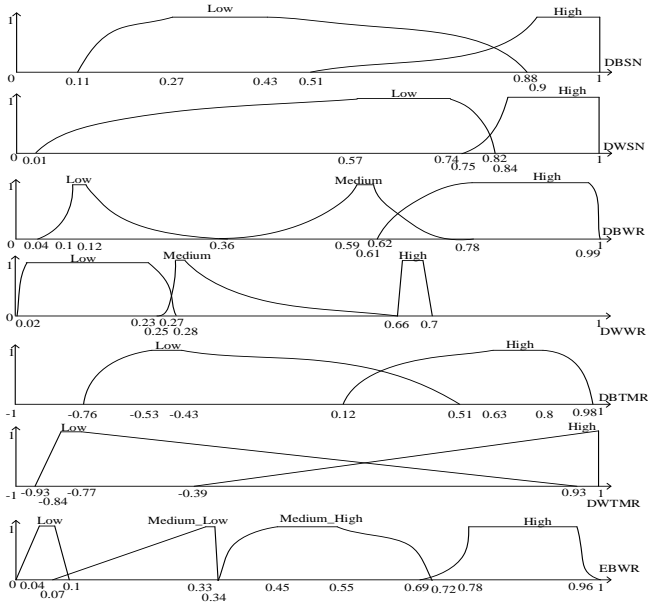

Fig. 7. Learned knowledge base of Black.

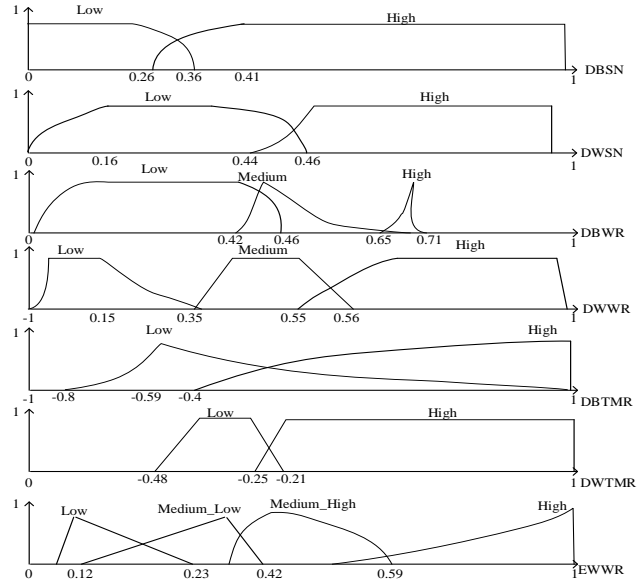

Fig. 8. Learned knowledge base of White.

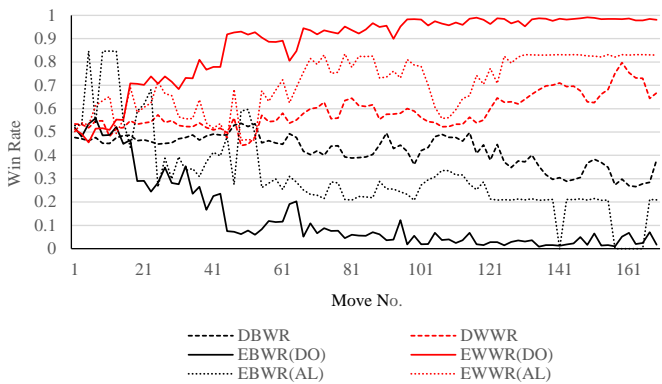

Fig. 9. Game 11: Black and White's win rate curves.

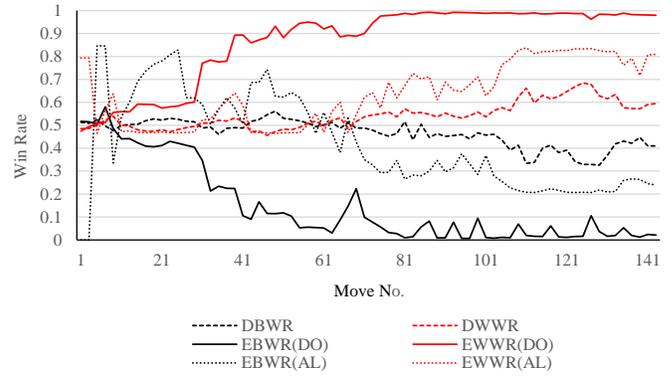

Fig. 10. Game 46: Black and White's win rate curves

## V. CONCLUSIONS

In this paper, we apply the evolutionary computing techniques to the human and machine cooperative learning on game of Go. A robot agent based on a genetic algorithm and fuzzy markup language is proposed. Additionally, the GFML-based robot agent is able to deploy on various robots, including the commercial robot Palro and Pepper, and the academic robot of Kubota Lab. in TMU for the human and machine cooperative learning system between Taiwan and Japan. The FML engineer and domain expert referred to the parameters of DDF and ELF OpenGo to construct the knowledge base and rule base of the robot agent. The data engineer and domain expert used the game data from Google AlphaGo Master sixty games as the training data and testing data of GFML. The experimental results showed that the robot agent can improve the knowledge base and rule based of the robot agent based on Darkforest and OpenGo AI bots with various simulation numbers. In the future, we will improve the GFML-based robot agent to be an easy to use machine learning tools on various robots for the human and machine co-learning system and extend this cooperative learning system to different countries.

### ACKNOWLEDGMENT

The authors would like to thank Dr. Yuandong Tian and Facebook AI Research (FAIR) ELF OpenGo/Darkforest team members for their open source and technical support. Additionally, we would like to thank Google DeepMind team to release "AlphaGo Master series: 60 online games." Finally, we would like to thank Yi-Hsiu Lee and Hirofumi Ohashi for their kind help with testing the stability of the cooperative learning system as well as thank Pin-Zhen Su for her kind help with organizing the predicted data of the sixty AlphaGo Master games.